\documentclass[conference]{IEEEtran}
\IEEEoverridecommandlockouts
\usepackage{cite}
\usepackage{amsmath,amssymb,amsfonts}
\usepackage{graphicx}
\usepackage{textcomp}
\usepackage{xcolor}
\usepackage[utf8]{inputenc}
\usepackage{amsmath}
\usepackage{array}
\usepackage{graphicx}
\usepackage[bookmarksopen,bookmarksdepth=2,breaklinks=true]{hyperref}
\usepackage{algorithm}
\usepackage{algpseudocode}
\usepackage{verbatim}
\usepackage{float}

\makeatletter
\newcommand{\linebreakand}{%
  \end{@IEEEauthorhalign}
  \hfill\mbox{}\par
  \mbox{}\hfill\begin{@IEEEauthorhalign}
}
\makeatother

\def\BibTeX{{\rm B\kern-.05em{\sc i\kern-.025em b}\kern-.08em
    T\kern-.1667em\lower.7ex\hbox{E}\kern-.125emX}}
\begin{document}

\title{PSL is Dead. Long Live PSL\\}

\author{\IEEEauthorblockN{Kevin Smith}
\IEEEauthorblockA{\textit{Rice University} \\
Houston, U.S.A. \\
kwsmith@rice.edu}
\and
\IEEEauthorblockN{Hai Lin}
\IEEEauthorblockA{\textit{Palo Alto Networks} \\
Santa Clara, U.S.A. \\
halin@paloaltonetworks.com}
\and
\IEEEauthorblockN{Praveen Tiwari}
\IEEEauthorblockA{\textit{Palo Alto Networks} \\
Santa Clara, U.S.A.\\
prtiwari@paloaltonetworks.com}
\linebreakand
\IEEEauthorblockN{Marjorie Sayer}
\IEEEauthorblockA{\textit{Palo Alto Networks} \\
Santa Clara, U.S.A. \\
msayer@paloaltonetworks.com}
\and
\IEEEauthorblockN{Claudionor Coelho}
\IEEEauthorblockA{\textit{Palo Alto Networks} \\
Santa Clara, U.S.A. \\
ccoelho@paloaltonetworks.com}
}

\maketitle

\begin{abstract}
Property Specification Language (PSL) is a form of temporal logic that has been mainly used in discrete domains (e.g. formal hardware verification). In this paper, we show that by merging machine learning techniques with PSL monitors, we can extend PSL to work on continuous domains. We apply this technique in machine learning-based anomaly detection to analyze scenarios of real-time streaming events from continuous variables in order to detect abnormal behaviors of a system. By using machine learning with formal models, we leverage the strengths of both machine learning methods and formal semantics of time. On one hand, machine learning techniques can produce distributions on continuous variables, where abnormalities can be captured as deviations from the distributions. On the other hand, formal methods can characterize discrete temporal behaviors and relations that cannot be easily learned by machine learning techniques. Interestingly, the anomalies detected by machine learning and the underlying time representation used are discrete events. We implemented a temporal monitoring package (TEF) that operates in conjunction with normal data science packages for anomaly detection machine learning systems, and we show that TEF can be used to perform accurate interpretation of temporal correlation between events.
\end{abstract}


\section{Introduction}

Property Specification Language (PSL) is a form of temporal logic that is designed to capture temporal relations between discrete variables over discrete time. Due to this nature, PSL has been mainly used in hardware design and verification since it was standardized by IEEE in 2004 \cite{Bloem2007, Boule2007, Javaheri2017}. There have been attempts to extend PSL to deal with continuous variables over continuous time \cite{Ge2021}. Due to its inherent limitation of expressibility, there have not been many successful applications.

In recent years, anomaly detection has been widely used in practice \cite{Chandola2009, Patcha2007}. There are many applications where real-time streaming events are monitored and analyzed in order to detect abnormal behaviors.
For example, if the amount of free memory of a computer is below a certain threshold, it can be considered as an anomaly. As another example, if there is an anomalous drop in purchase of a product in an online store, it is possible that the product is out of stock, which needs attention.
The state-of-the-art technique for anomaly detection is machine learning \cite{Cui2022, Jain2022, Kasim2020, Li2021, Ma2022, Pei2022}. Machine learning techniques learn distributions on continuous variables. Anomaly events can be captured as deviations from established patterns (distributions).
However, there are certain temporal behaviors and relations that cannot be easily learned by machine learning techniques, but can be easily characterized by formal languages such as PSL.

\begin{figure}
\begin{center}
    \includegraphics[scale=0.4]{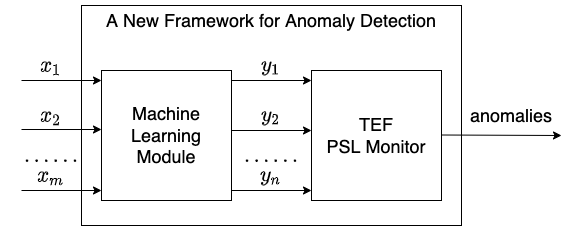}
\end{center}

\caption{A New Framework for Anomaly Detection}
    \label{img:framework}
\end{figure}

In this paper, we propose a new framework called TEmporal Filtering (TEF) for anomaly detection (Fig. \ref{img:framework}). The idea is to merge machine learning with PSL monitors. The machine learning module takes as input a number of continuous variables $x_1, x_2, \ldots\ldots, x_m$, and outputs some discrete events $y_1, y_2, \ldots\ldots, y_n$, which become the input of the PSL monitor. The PSL monitor encodes a user-defined temporal relation, which filters the output from the machine learning module. In this new framework, machine learning techniques extend the capability of PSL by discretizing continuous time and events; the PSL monitor refines the results produced by the machine learning module. This combination of machine learning and formal methods yields a whole that is greater than the sum of its parts.

The rest of this paper is organized as follows. In Section \ref{sec:anomaly}, we give a brief introduction to anomaly detection. Section \ref{sec:overview} discusses the overall architecture of TEF. In Section \ref{sec:implementation}, we describe how TEF is implemented. Section \ref{sec:casestudies} illustrates how TEF can be used to capture temporal relations. Section \ref{sec:future} summarizes the conclusions of the paper and future work. 

\section{Anomaly Detection} \label{sec:anomaly}

Anomaly detection is the process of identifying events that deviate from established patterns. Anomaly detection is widely used in many applications such as detecting cyber intrusions, credit card fraud, and health monitoring. In many applications, input variables have continuous values that vary with time, such as temperature. A time series anomaly detection model learns baseline behavior from training data and predicts a discrete set of anomalies. In time series anomaly detection the discretizing mechanism is simply that input timestamps are taken from discrete measurements, and there will always be a minimum nonzero granularity of inputs. These models provide a rich set of examples where a continuous valued problem maps to a discrete space, where formal methods can be applied. 

Anomaly detection modeling faces two major challenges. One, unbalanced data sets: anomalies are rare, and thus data sets will have few examples of anomalies which the model can learn. Two, characterization of anomalies: different types of models detect different types of anomalies. Level based methods find metric outliers. Distribution based methods find anomalies in distributions. Features might not contain the signal needed to detect anomalies. Being able to characterize anomalies helps, but unseen anomalies cannot be characterized. In such cases, model selection is difficult. The next sections describe some common types of anomaly detection models. 

\subsection{Level-Based Anomaly Detection}
\label{subsec:level-based}In these types of models, anomalies in continuous data are defined as values beyond a specific threshold. The threshold level is typically calibrated to the expected fraction of anomalies to be detected. A level that is too low results in false negatives, and one that is high results in false positives. Often an immediate limitation of level based models is they do not account for the frequency of threshold crossings. But threshold crossings and the timestamps they occur do make up a discrete event space for study. 

\subsection{Distribution-Based Anomaly Detection}
\label{subsec:distribution-based} Distribution based models learn the statistical distribution of data in a baseline or normal state. Anomalies are categorized as events whose predicted probabilities are lower than a learned threshold. These models can learn more sophisticated behavior than level-based models - in particular, nonlinear decision boundaries between anomalous and normal events can be learned. The challenge of tuning the model remains: both the parameters that affect the machine learning of the model, and the final tuning of the anomaly probability thresholds. These challenges are documented in \cite{Kim2019}. 

\subsection{Forecasting Error Methods}
\label{subsec:forecast-based}A wide range of time series forecasting methods, such as ARIMA, can be used to predict probable events from past data. Predictions from a recent past period can be compared with actual data values to determine if the actual values are anomalous. The comparison of prediction and truth can in turn be level based or distribution based. 

\subsection{Template-based Anomaly Detection}
\label{subsec:template-based}If anomalies follow templates of behavior, established rules of feature interaction and evolution through time, it's reasonable to hope that a machine learning model will learn the rules. The success of machine learning in many applications has led to extensive efforts in applying machine learning models to anomaly detection. Given enough features, enough data, enough compute power, the reasoning goes, a machine learning model will learn all of the intricate influences that distinguish anomalous behavior from normal. First, we demonstrate that there are cases where enough data is theoretically impossible. For instance, in a time stream of data, the event consisting of repeated events a followed by an event b are not possible to learn from a finite dataset. Second, while it may be possible to learn an underlying template for anomalies, the amount of data and compute resources required for sufficiently accurate results might be prohibitive. 

\subsection{A Hybrid Method}
\label{subsec:proposed-ML-TEF}Because anomaly detection models create discrete sets of events they naturally can be described using PSL. Machine learning can make PSL relevant in continuous applications. Because PSL can easily characterize infinite sets such as the "arbitrary stream of a followed by b" example as a simple PSL expression: $a[+];b$, machine learning anomaly detection can be enhanced. We aim to show in this paper that time series anomaly detection together with TEF can improve overall model performance. 



\section{TEF Overview} \label{sec:overview}

TEF implements a subset of PSL. PSL is an extension of linear temporal logic and adds a number of operators to express temporal constraints. In particular, PSL makes use of Sequential Extended Regular Expressions (SEREs), defined below. If a PSL formula is composed entirely of SEREs, it is said to be written in SERE-style PSL. TEF implements most of SERE-style PSL.

Propositional formulas (i.e., boolean variables closed under conjunction, disjunction, and negation) are the atoms of SEREs. SEREs are SERE atoms closed under the SERE operators, which are analogous to the operators of regular expressions and also include supplemental operators representing useful syntactic sugar. Just as regular expressions are used to match strings, SEREs are used to match traces, that is, sequences of truth assignments. A SERE atom matches a truth assignment when the truth assignment makes the atom true. The semantics of compound SEREs are shown in Fig. \ref{fig:mainoperators} and Fig. \ref{fig:syntacticsugar}.

One of the advantages in using a temporal logic to specify properties is that logics are declarative. The result can be precisely described without the use of control flow or statements that modify a program's state. This allows those without a background in software engineering to write useful properties. However, the operators of linear temporal logic, although conceptually simple, are not trivial to use in practice, and most people are unfamiliar with them. An advantage of SERE-style PSL is that the functionality of LTL operators is subsumed by the SERE operators and so familiarity with LTL is not a requirement to writing properties. The close resemblance between SEREs and standard regular expressions make the former exceptionally easy to learn if one is familiar with the latter. And regular expressions are common currency not just in software engineering, but in data analysis and related fields as well. This makes SERE-style PSL easily accessible to those with a wide variety of backgrounds.

So SERE-style PSL is a natural choice for TEF, which aims to provide a simple and accessible way to express and evaluate temporal constraints. In fact, TEF extends SERE-style PSL in an intuitive and useful way by allowing the use of Boolean expressions wherever Boolean variables may occur in SEREs. Another natural choice is the decision to package TEF as an extension to Pandas, which is also common currency in the Python world.

TEF makes use of Python's regex engine, which takes as input a regex pattern (regular expression) and a string. The regex engine reports all segments of the string that match the regex pattern. Python's regex engine is highly optimized and efficient.
Fig. \ref{fig:mainoperators} lists the main temporal operators, which TEF implements, and compares them with Python's regex patterns. In this table, $r$ values are SEREs, $s$ values are sequences of rows in a DataFrame, $r'$ values are regex patterns, $s'$ values are strings.
Readers are referred to \cite{Pill2008} for the formal semantics of SERE-style expressions.

\begin{figure}[h]
\centering
  \includegraphics[scale=0.41]{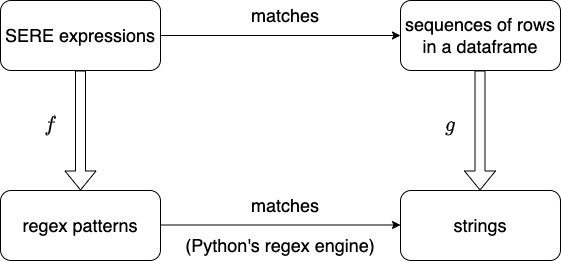}
  \caption{How TEF is implemented}
  \label{img:reduction}
\end{figure}

At a high level, TEF takes advantage of the highly efficient regex engine and the similarity between SERE expressions and regex patterns (shown in Fig. \ref{fig:mainoperators}).
TEF reduces the problem of checking if a SERE expression matches rows of a data frame to checking if a regex expression matches a string (Fig. \ref{img:reduction}).
The reduction is based on the following observation.
Let $\mathcal{E}$ be the set of all SERE expressions, $\mathcal{R}$ be the set of sequences of rows in a data frame, $P$ be the set of all regex patterns, and $S$ be the set of strings. There exist two functions $f: \mathcal{E} \rightarrow \mathcal{P}$ and $g: \mathcal{R} \rightarrow \mathcal{S}$ s.t. $\forall e\!:\!\mathcal{E}, \;\; r\!:\!\mathcal{R}. \;\; matches(e, r) \leftrightarrow matches(f(e), g(r)) $.
Intuitively, Fig. \ref{fig:mainoperators} justifies this observation.

TEF also implements some additional operators as syntatic sugar (shown in Fig. \ref{fig:syntacticsugar}).
TEF also adds the ``[]'' operator for use in Boolean expressions, which allows further flexibility in the expression of temporal relations. It functions in the following way. Suppose we have a data frame with a column $c$. Then $c[-1]$ refers to the value at the previous row in column $c$. Also, $c[1]$ refers to the value at the next row in column $c$. In general, if $i$ is the index of the current row and $j$ is an integer, then $c[j]$ refers to the value at row $i + j$ of column $c$. As syntactic sugar, $c$ may be used as an abbreviation for $c[0]$. An illustrative example of a property making use of the ``[]'' operator is $(c>c[-1])[*5]$, which matches all segments of the trace in which the value at $c$ increases five times consecutively.

\begin{figure*}

\resizebox{\textwidth}{!}{
\begin{tabular}{ |c|c|c|c| }

 \hline
 \text{SERE Syntax} & \text{Meaning} & \text{Regex Syntax} & \text{Meaning} \\
 \hline\hline
 $r_1 ; r_2$ & $\text{matches }s \text{ if } s=s_1s_2, r_1 \text{ matches } s_1 \text{  and } r_2 \text{ matches } s_2$ & $r_1'r'_2$ & $\text{matches }s' \text{ if } s'=s'_1s'_2, r'_1 \text{ matches } s'_1 \text{  and } r'_2 \text{ matches } s'_2$ \\ 
 \hline
 $r_1 | r_2$ & $\text{matches }s \text{ if } r_1 \text{ matches } s \text{  or } r_2 \text{ matches } s$ & $r'_1 | r'_2$ & $\text{matches }s' \text{ if } r'_1 \text{ matches } s' \text{  or } r'_2 \text{ matches } s'$ \\ 
 \hline
 $r_1 \& r_2$ & $\text{matches }s \text{ if } r_1 \text{ matches } s \text{  and } r_2 \text{ matches } s$ & $?= r'$ & $\text{matches }s' \text{ if } r' \text{ matches } s', but it does not consume any r'$\\ 
 \hline
 $r [*]$ & $\text{matches } s \text{ if } 0 \text{ or more concatenations of } r \text{ matches } s$ & $r' *$ & $\text{matches } s' \text{ if } 0 \text{ or more concatenations of } r' \text{ matches } s'$ \\ 
 \hline
 $r [+]$ & $\text{matches } s \text{ if } 1 \text{ or more concatenations of } r \text{ matches } s$ & $r' +$ & $\text{matches } s' \text{ if } 1 \text{ or more concatenations of } r' \text{ matches } s'$ \\ 
 \hline
 $r[*n]$ & $\text{matches } s \text{ if } n \text{ concatenations of } r \text{ matches } s$ &  $r'\{n\}$ & $\text{matches } s' \text{ if } n \text{ concatenations of } r' \text{ matches } s'$ \\ 
 \hline
 $r[*n..m]$ & $\text{matches } s \text{ if between } n \text{ and } m \text{ concatenations of } r \text{ matches } s$ & $r'\{n,m\}$ & $\text{matches } s' \text{ if between } n \text{ and } m \text{ concatenations of } r' \text{ matches } s'$ \\ 
 \hline
 $r[*n..]$ & $\text{matches } s \text{ if } n \text{ or more concatenations of } r \text{ matches } s$ & $r'\{n,\}$ & $\text{matches } s' \text{ if } n \text{ or more concatenations of } r' \text{ matches } s'$ \\ 
 \hline
 $r[*..m]$ & $\text{matches } s \text{ if } m \text{ or fewer concatenations of } r \text{ matches } s$ & $r\{0, m\}$ & $\text{matches } s' \text{ if } m \text{ or fewer concatenations of } r' \text{ matches } s'$ \\ 
 \hline
 
\end{tabular}

}
\caption{SERE expressions and Python's regex patterns}
    \label{fig:mainoperators}
\end{figure*}

\begin{figure}
\centering
\resizebox{0.26\textwidth}{!}{

$
\begin{array}{ |c|c| } 

\hline
 \text{Syntax} & \text{Intuitive Meaning} \\ 
 \hline\hline
[+] & \text{True}[+] \\ 
 \hline
 [*] & \text{True}[*] \\ 
 \hline
 [*n] & \text{True}[*n..] \\ 
 \hline
 [*n..m] & \text{True}[*n..m] \\
 \hline
 [*n..] & \text{True}[*n..] \\ 
 \hline
 [*..m] & \text{True}[*..m] \\ 
 \hline
 r[->] & !r[*];r \\ 
 \hline
 r[-> n] & !r[*];r[*n] \\ 
 \hline
 r[-> n..m] & !r[*];r[*n..m] \\ 
 \hline
 r[=n] & (!r[*] ; r)[*n] ; !r[*] \\ 
 \hline
 r[=n..m] & (!r[*] ; r)[*n..m] ; !r[*] \\ 
 \hline
 r[=n..] & (!r[*] ; r)[*n..] ; !r[*] \\ 
 \hline
 r[=..m] & (!r[*]; r)[*..m] ; !r[*] \\ 
 \hline
\end{array}
$
}
\caption{Additional SERE-style operators}
    \label{fig:syntacticsugar}
\end{figure}

\begin{algorithm}
\caption{An algorithm for converting a SERE expression $e$ to a regex expression}\label{alg:f}

\begin{algorithmic}[1]
\Function{psl\_to\_regex}{$e$}
  \If {$e$ is a Boolean expression $b$}
    \State \Return the disjunction of truth assignments that make $b$ true, in string form.
  \ElsIf{$e$ is $r_1;r_2$}
    \State $s_1 \gets$ \Call{psl\_to\_regex}{$r_1$}
    \State $s_2 \gets$ \Call{psl\_to\_regex}{$r_2$}
    \State \Return `(' + $s_1 + s_2$ + `)'     
  \ElsIf{$e$ is $r_1 | r_2$}
    \State $s_1 \gets$ \Call{psl\_to\_regex}{$r_1$}
    \State $s_2 \gets$ \Call{psl\_to\_regex}{$r_2$}
    \State \Return `(' + $s_1 +$ `$|$' $+ s_2$ + `)'
  \ElsIf{$e$ is $r_1 \& r_2$}
    \State $s_1 \gets$ \Call{psl\_to\_regex}{$r_1$}
    \State $s_2 \gets$ \Call{psl\_to\_regex}{$r_2$}
    \State left = '(' + '(?=' + $s_1$ + ')' + $s_2$ + ')'
    \State right = '(' + '(?=' + $s_2$ + ')' + $s_1$ + ')'
    \State \Return '(' + left + '$|$' + right + ')'
  \ElsIf{$e$ is $r[*]$}
    \State $s \gets$ \Call{psl\_to\_regex}{$r$}
    \State \Return `(' + $s$ + `*' + `)'
  \ElsIf{$e$ is $r[+]$}
    \State $s \gets$ \Call{psl\_to\_regex}{$r$}
    \State \Return `(' + $s$ + `+' + `)'
  \ElsIf{$e$ is $r[*n]$}
    \State $s \gets$ \Call{psl\_to\_regex}{$r$}
    \State \Return '((' + s + ')' + '\{' + n + '\}' + ')'
  \ElsIf{$e$ is $r[*n..m]$}
    \State $s \gets$ \Call{psl\_to\_regex}{$r$}
    \State \Return '((' + s + ')' + '\{' + n + `,' + m + '\}' + ')'
  \ElsIf{$e$ is $r[*n..]$}
    \State $s \gets$ \Call{psl\_to\_regex}{$r$}
    \State \Return '((' + s + ')' + '\{' + n + `,' + '\}' + ')'
  \ElsIf{$e$ is $r[*..m]$}
    \State $s \gets$ \Call{psl\_to\_regex}{$r$}
    \State \Return '((' + s + ')' + '\{0' + `,' + m + '\}' + ')'
  \EndIf
\EndFunction

\end{algorithmic}
\end{algorithm}

\section{TEF Implementation} \label{sec:implementation}

TEF shadows the Pandas DataFrame eval method. It takes as input a SERE-style PSL property in the form of a string, and it matches it against the DataFrame, which is interpreted as a trace.
As we discussed in Section \ref{sec:overview}, TEF reduces the problem of checking if a SERE expression matches rows of a data frame to checking if a Python regex pattern matches a string.
In order to make the reduction work, we just need to build two functions: (1) $f$ which converts a SERE expression to a regex pattern. (2) $g$ which converts a data frame to a string.

To build the function $f$,
we use a simple recursion based on how the SERE expression is constructed.
The simplest SERE expression is a Boolean expression. To convert a Boolean expression to a regex pattern, we rewrite the expression in the form of the disjunction of its set of satisfying truth assignments, in string form. As a result, the length of the pattern is exponential in the number of distinct boolean expressions used in the property. In practice we have found that many problems need make use of only a few expressions. Often only a single expression is used to express a useful property.
To convert more complicated SERE expressions involving SERE operators, the individual pieces are converted recursively, then the results are combined based on the SERE operator. For example, suppose that we want to convert $r_1; r_2$. We first recursively convert $r_1$ and $r_2$, get two regex strings, and then concatenate the two regex strings.
The details of converting a SERE expression into a Python regex pattern is shown in Algorithm \ref{alg:f}.

To build the function $g$, TEF does two steps:

\begin{itemize}
    \item \textbf{Step 1} (Booleanize a data frame) TEF identifies all the Boolean expressions used in the property, and evaluates them with respect to each row in the data frame. It keeps track of the results in a new data frame, in which the results of each Boolean expression are kept in new columns. The details of Booleanizing a data frame is shown in Algorithm \ref{alg:g}.
 
    \item \textbf{Step 2} (convert to a string) The Booleanized DataFrame is converted to a string by using `,' to seperate columns and using `()' to group characters within the same rows.
\end{itemize}

After converting a SERE expression into a regex pattern and converting a data frame into a string,
Python's regex engine is used to find matches. Results are returned in the form of a list of index pairs indicating where in the trace the property is satisfied.

\begin{algorithm}
\caption{An algorithm for Booleanizing a data frame $df$ w.r.t. a compound Boolean expression $e$}\label{alg:g}
\begin{algorithmic}[1]
\Function{Booleanize\_dataframe}{$df, e$}
  \State Initialize $b\_df$ to be an empty data frame.
  \ForAll{row $r$ in $df$}
    \State Initialize $r'$ to be an empty row.
    \ForAll{atom $a$ in $e$}
        \State Evaluate $a$ w.r.t. $r$, add the result to $r'$
    \EndFor
    \State Add $r'$ to $b\_df$
  \EndFor
  \State \Return $b\_df$
\EndFunction
\end{algorithmic}
\end{algorithm}



\section{Case Studies} \label{sec:casestudies}

\subsection{Analyzing Weather Patterns using TEF}

\begin{figure}[ht]
\centering
\begin{minipage}{.24\textwidth}
  \centering
  \includegraphics[width=0.9\linewidth, height=12cm]{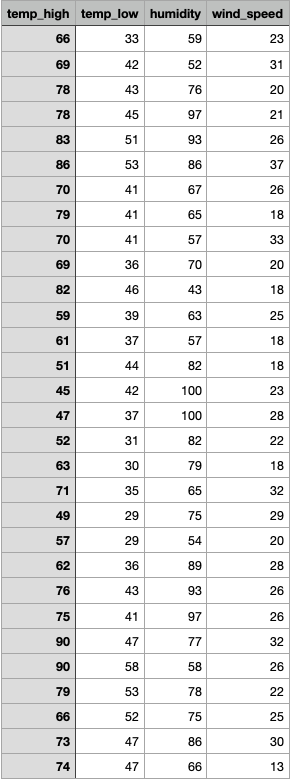}
  \caption{Data Frame: \textit{df}}
    \label{img:temperature}
\end{minipage}%
\begin{minipage}{.24\textwidth}
  \centering
  \includegraphics[width=0.8\linewidth, height=12cm]{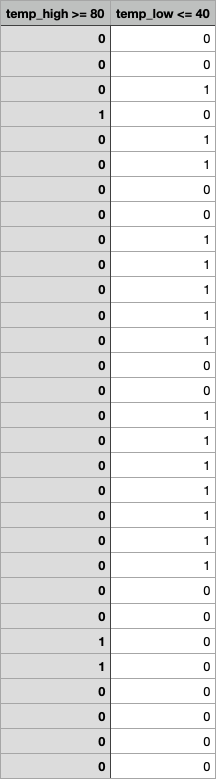}
  \caption{Booleanized Data Frame: \textit{b\_df}}
    \label{img:bool_temperature}
\end{minipage}
\end{figure}

Figure \ref{img:temperature} shows a data frame containing the weather information in Amarillo, TX, in April 2021. The weather data is taken from \cite{weather}.
We make the following queries to the data frame. Pandas DataFrame eval method can handle queries involving only one row. If the query refers to multiple rows, we need to use TEF eval method.

\begin{figure}[ht]
\centering
  \includegraphics[scale=0.24]{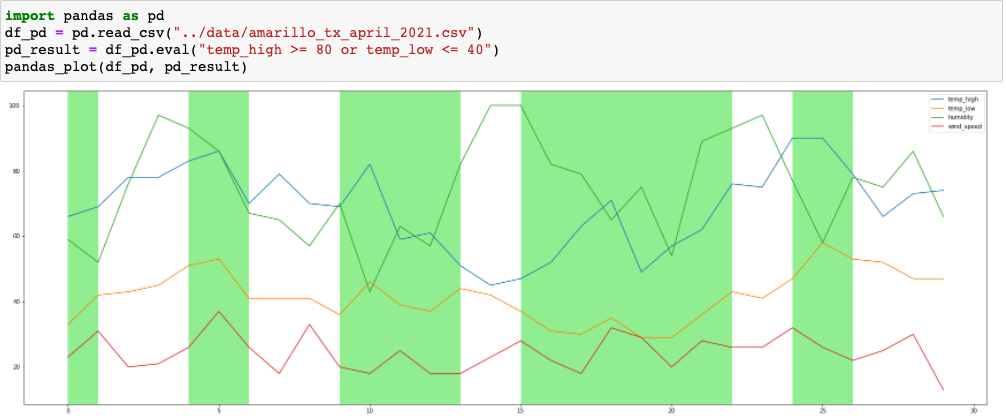}
  \caption{Query using Pandas DataFrame eval method}
  \label{img:example1}
\end{figure}

1. Find all individual days, when the temperature is either too hot (temp\_high $\geq$ 80) or too cold(temp\_low $\leq$ 40). We can use Pandas DataFrame eval method, which returns the set of individual rows, where the constraint is satisfied (Fig. \ref{img:example1}). 

2. Find two consecutive days where a hot day (temp\_high $\geq$ 80) is followed by a cold day (temp\_low $\leq$ 40). The Pandas DataFrame eval method cannot handle this query, since it cannot reason about temporal relations involving multiple rows of a data frame. We use TEF eval method instead. TEF does the following 3 steps to evaluate ``temp\_high $\geq$ 80; temp\_low $\leq$ 40'' against the data frame $df$ (Fig. \ref{img:temperature}).

\begin{figure}[ht]
\centering
  \includegraphics[scale=0.24]{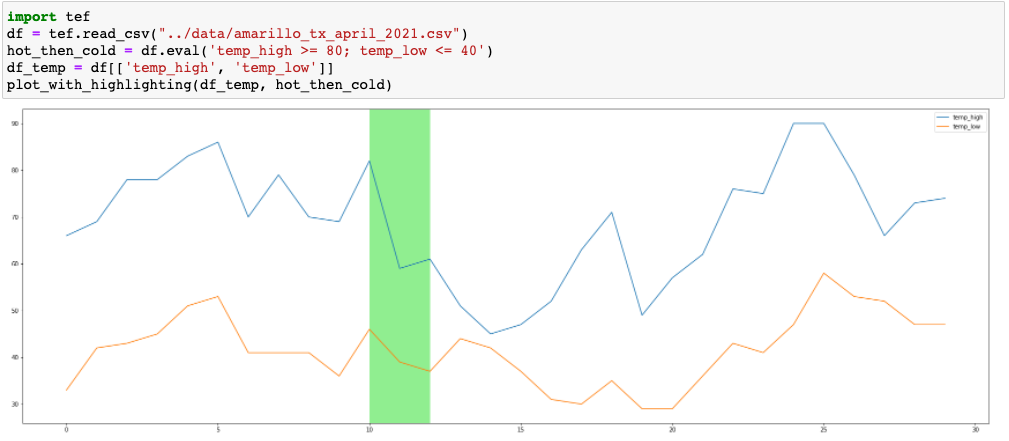}
  \caption{Query using TEF eval method}
  \label{img:example2}
\end{figure}

\begin{itemize}
    \item \textbf{Step 1:} As discussed in Section \ref{sec:implementation}, TEF identifies all the Boolean expressions used in the property, and evaluates them with respect to each
row in \textit{df}. This particular query contains two Boolean expressions ``high\_temperature $\geq$ 80'' and ``low\_temperature $\leq$ 40''. The Booleanized data frame \textit{b\_df} (Fig. \ref{img:bool_temperature}) has two columns and has the same number of rows as \textit{df}. \textit{b\_df} is transformed into the following string by using `,' to seperate columns and using `()' to group characters within the same rows.

\medskip

(0,1)(0,0)(0,0)(0,0)(1,0)(1,0)(0,0)(0,0)(0,0)(0,1)(1,0)

(0,1)(0,1)(0,0)(0,0)(0,1)(0,1)(0,1)(0,1)(0,1)(0,1)(0,1)

(0,0)(0,0)(1,0)(1,0)(0,0)(0,0)(0,0)(0,0)

\medskip

\item \textbf{Step 2:} The PSL property is compiled to a Python regex pattern: $(\backslash \left(1,0\backslash\right) | \backslash\left( 1,1\backslash\right)) (\backslash\left(0,1\backslash\right) | \backslash\left(1,1\backslash\right))$.

\item \textbf{Step 3:} Python’s regex engine is used to find
matches.
\end{itemize}

The result is shown in Fig. \ref{img:example2}.

\begin{figure}[ht]
\centering
  \includegraphics[scale=0.24]{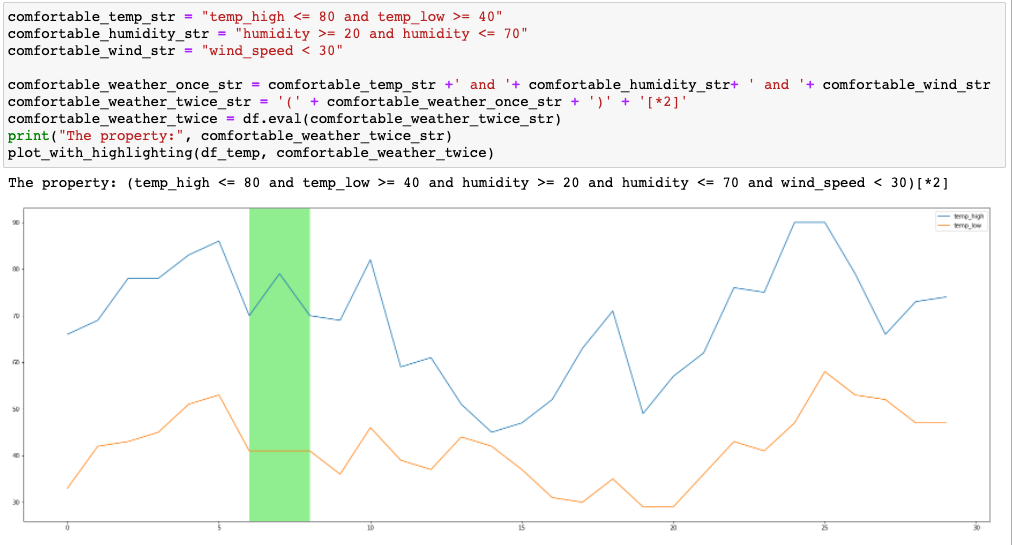}
  \caption{Query using TEF eval method}
  \label{img:example3}
\end{figure}

3. Find two consecutive days when (1) temp\_high $\leq$ 80 and temp\_low $\geq$ 40 (2) humidity $\geq$ 20 and humidity $\leq$ 70 (3) wind\_speed $<$ 30. Again, this is a temporal relation involving multiple rows of the data frame. We should use TEF eval method. The result is shown in Fig. \ref{img:example3}.

\subsection{Analyzing Dow Jones Industrial Average (DJIA) index using TEF}

In this section, we show how TEF can be used to analyze patterns in Dow Jones Industrial Average (DJIA) index. This dataset contains data from 01/01/1980 to 12/31/2012, and is taken from \cite{DJIA}. All experiments are done on a Macbook, with 2.6 GHz 6-Core Intel Core i7 and 16 GB 2667 MHz DDR4.

\begin{figure}[ht]
\centering
  \includegraphics[scale=0.24]{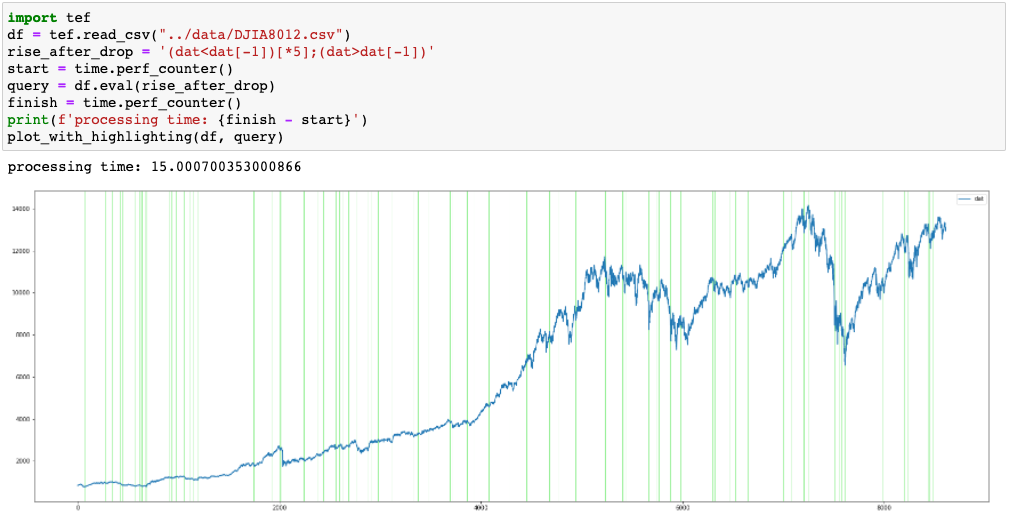}
  \caption{Rise\_after\_Drop}
  \label{img:djia1}
\end{figure}

1. [\textbf{Rise\_after\_Drop}] Find all periods where the index has been dropping for 5 consecutive days and rises on the next day. The result is shown in Fig. \ref{img:djia1}.

\begin{figure}[ht]
\centering
  \includegraphics[scale=0.24]{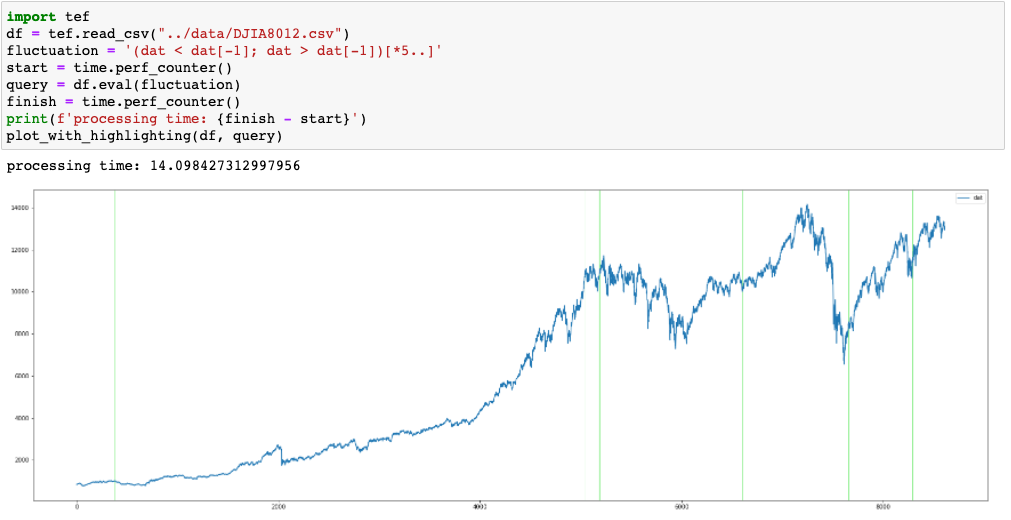}
  \caption{Flunctuation}
  \label{img:djia2}
\end{figure}

2. [\textbf{Fluctuation}] Find all periods where the index decreases on one day and increases on the next day, and this pattern repeats for at least 5 times. The result is shown in Fig. \ref{img:djia2}.

\begin{figure}[ht]
\centering
  \includegraphics[scale=0.24]{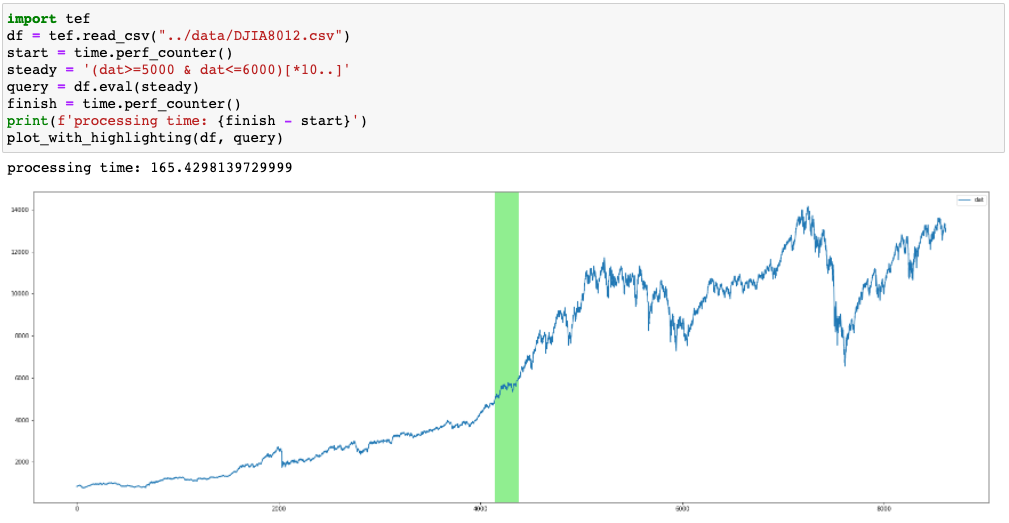}
  \caption{Steady}
  \label{img:djia3}
\end{figure}

3. [\textbf{Steady}] Find all periods where the index stays between 5000 and 6000 for at least 10 days. The result is shown in Fig. \ref{img:djia3}.

\begin{figure}[ht]
\centering
  \includegraphics[scale=0.24]{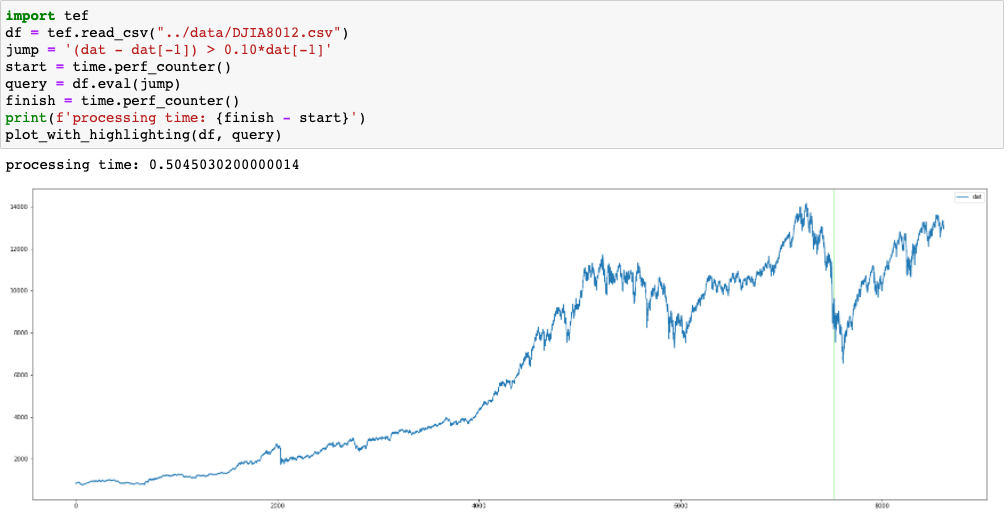}
  \caption{Jump}
  \label{img:djia4}
\end{figure}

4. [\textbf{Jump}] Find all periods where the index increases by at least 10\% compared to the previous day. The result is shown in Fig. \ref{img:djia4}.

\subsection{Isolation Forest Anomaly Detection Models and TEF}
Using examples from publicly available data sets, we show cases where false positives from a trained isolation forest model can be filtered by rules. The data is taken from \cite{UCR-AD}. The entire dataset collection consists of 250 datasets from domains such as medical monitoring, motion sensors, and weather. We identified 130 examples where the model output could be improved by a rule based filter. 

\begin{figure}[ht]
\centering
  \includegraphics[scale=0.4]{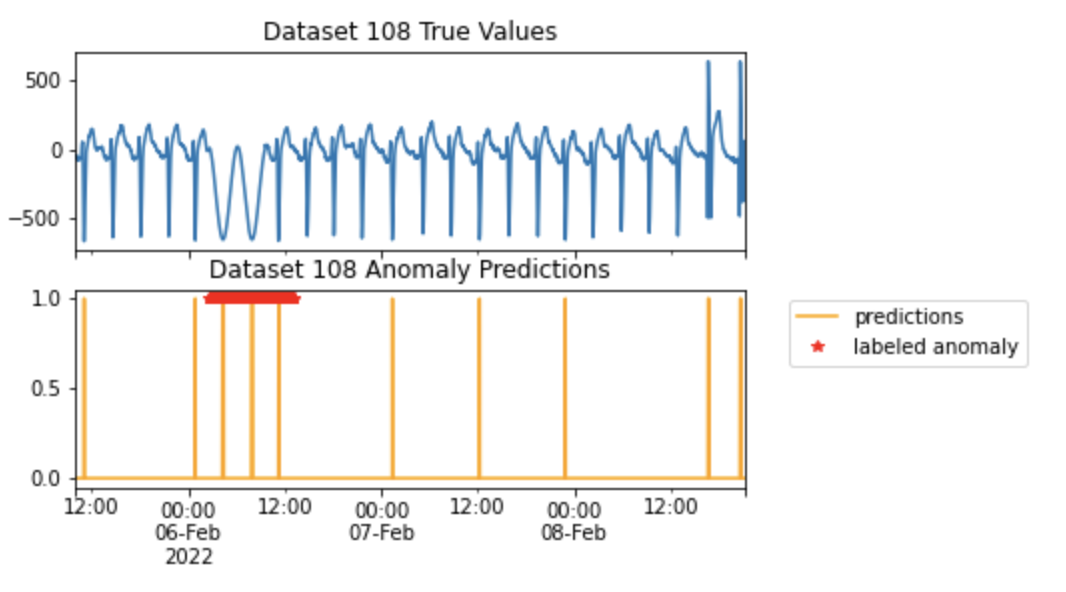}
  \caption{Identify cluster of predictions}
  \label{img:cluster1}
\end{figure}

\begin{figure}[ht]
\centering
  \includegraphics[scale=0.4]{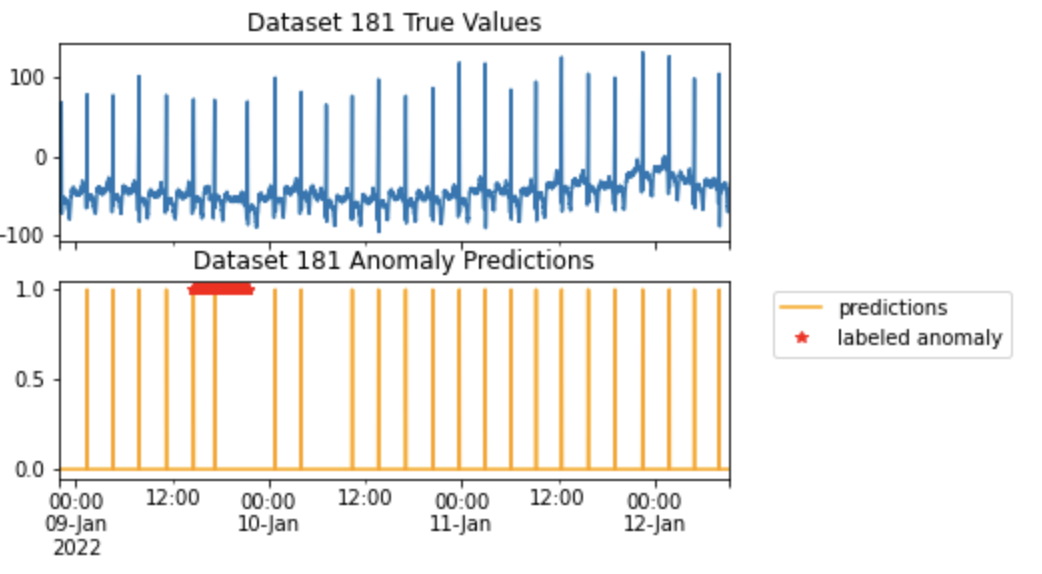}
  \caption{Identify by gap}
  \label{img:gap}
\end{figure}

In Fig. \ref{img:cluster1} the anomaly is difficult to detect as it has similar amplitude and frequency to the baseline pattern. But the detected anomalies are clustered in time and anomalies that are farther apart could be ruled out by a simple rule filter. We can use TEF to specify that within a cluster, the total number of anomalies needs to reach a certain threshold. For example, $cluster_{2, 5}$ = ``$anomaly[-2] + anomaly[-1] + anomaly + anomaly[1] + anomaly[2] \geq 2$'' specifies that at least 2 anomalies need to occur within a cluster of width 5. The width of the cluster and the threshold can be adjusted for different applications.
Fig. \ref{img:gap} is an example where the anomaly can be detected by a gap that follows detected anomalies. In order to specify that a gap of width $x$ needs to follow a detected anomaly, we can use: ``$[*]; anomaly; !anomaly[*x]$''. The original data, shown in the upper graph, has an anomaly that is difficult to characterize. 

\begin{figure}[ht]
\centering
  \includegraphics[scale=0.4]{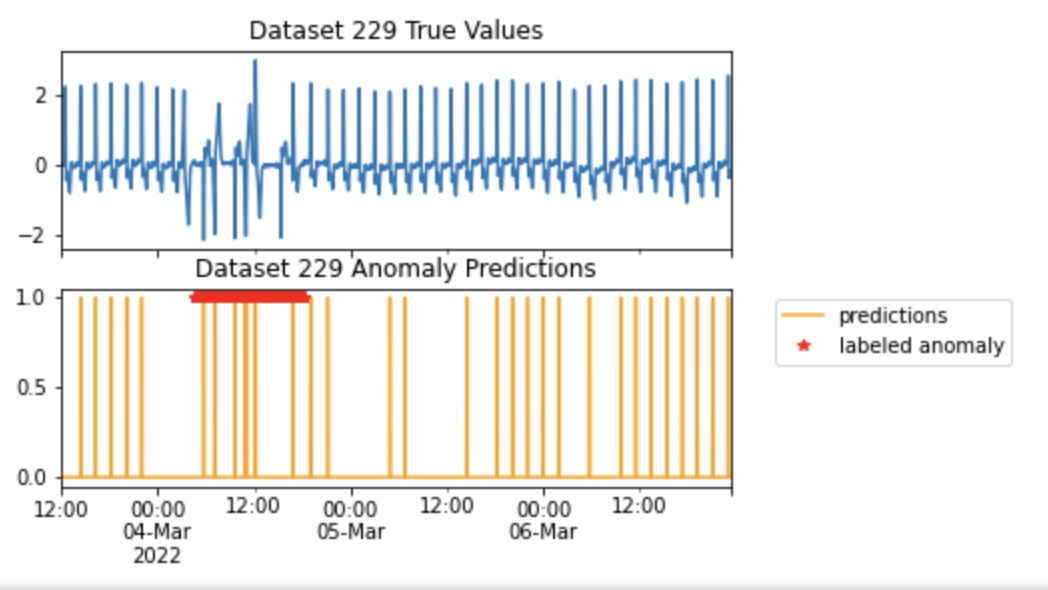}
  \caption{Identify by gap}
  \label{img:pattern}
\end{figure}

As a final example, Fig. \ref{img:pattern} shows a case where a suitable filtering rule would take into account both cluster and gap behavior. ``$cluster_{2, 5}; !anomaly[*5]; cluster_{2, 5}$'' specifies a sequence of events: at least 2 anomalies occur within a cluster of width 5, followed by a gap of width 5, followed by at least 2 anomalies occurring within a cluster of width 5. Again, the width of the cluster and the width of the gap can be adjusted for different applications.

\section{Conclusions and Future Work} \label{sec:future}

For many years, the use of PSL has been limited to hardware design and verification, since PSL is designed to deal with discrete variables over discrete time. In this paper, we propose that PSL can be used in continuous domain for anomaly detection. The idea is to merge PSL with machine learning and use a hybrid framework for anomaly detection. This hybrid framework consists of a machine learning module and a PSL monitor. The machine learning module outputs anomalies in the form of discrete time and events. The PSL monitor further refines the output of the machine learning module by checking if some user-defined temporal relation is satisfied. We implemented a temporal monitoring package (TEF), and we show that TEF can be used to perform accurate interpretation of temporal correlation between events.

For future work we are experimenting with the following cases: multivariate anomaly detection problems, and TEF in conjunction with ensemble models. With TEF, we have the opportunity to transform anomaly characterization rules into data, and develop models from there.


\end{document}